\documentclass[12pt,a4paper]{article}

\usepackage{essv}
\usepackage{float}
\usepackage{amsmath}
\usepackage{amssymb}
\usepackage{multirow}
\usepackage{makecell}
\usepackage[utf8]{inputenc}
\usepackage{xcolor}
\usepackage{footnote}
\usepackage{enumitem}
\usepackage{tikz}
\usetikzlibrary{positioning,decorations.pathreplacing,shapes,arrows,fit}
\usepackage[justification=centering]{caption}

\DeclareMathOperator*{\argmax}{arg\,max}

\title{IMS-Speech: A Speech to Text Tool}
\author{Pavel Denisov, Ngoc Thang Vu}
\affil{Institute for Natural Language Processing (IMS), University of Stuttgart}
\email{\{pavel.denisov|thang.vu\}@ims.uni-stuttgart.de}

\begin{document}

\selectlanguage{english}

\maketitle

\begin{abstract}
We present the IMS-Speech, a web based tool for German and English speech transcription aiming to facilitate research in various disciplines which require accesses to lexical information in spoken language materials. 
This tool is based on modern open source software stack, advanced speech recognition methods and public data resources and is freely available for academic researchers.
The utilized models are built to be generic in order to provide transcriptions of competitive accuracy on a diverse set of tasks and conditions.
\end{abstract}

\section{Introduction}

There is a considerable amount of spoken language materials in form of audio recordings, which researchers in e.g. humanities and social science could incorporate into their studies.
However, to be able to access to their content, one needs to automatically transcribe these recordings. 
While all needed resources for building of an automatic speech recognition (ASR) system are typically available for academic usage, their utilization requires specialized knowledge and technical experience \cite{povey2011kaldi}, \cite{watanabe2018espnet}.
Therefore, in order to provide people easy accesses to information in spoken language materials, a speech to text tool with a user interface should be helpful.

This paper presents the IMS-Speech\footnote{\url{http://www.ims.uni-stuttgart.de/forschung/ressourcen/werkzeuge/IMS-Speech.html}},
a web based tool for German and English speech transcription aiming to facilitate research in various disciplines. 
We are willing to provide a speech transcription service with an intuitive web interface accessible with a wide range of computing devices and to people with various backgrounds.
The service is based on modern open source software stack, advanced speech recognition methods and public data resources and is freely available for academic researchers.
The utilized models are built to be generic in order to provide transcriptions of competitive accuracy on a diverse set of tasks and conditions.
In addition to that, they can serve as a strong base for customized task specific applications.
 
\section{System description}

In order to produce a meaningful transcription for the most
of recordings that might be uploaded by users,
two tasks must be performed for every recording sequentially.
First, a recording must be split to segments
not exceeding some short duration and corresponding to speech intervals.
Second, actual ASR must be performed over each speech segment
for finding the most probable sequence of words being said
in the segment and thus constructing final transcription.

\subsection{Speech Segmentation}

Speech segmentation is performed with a speech activity detection (SAD)
system based on Time-Delay Neural Network (TDNN) \cite{waibel1990phoneme}
with statistics pooling \cite{ghahremani2016acoustic} for long-context information.
TDNN is trained to estimate probability of 3 classes,
\textit{Silence}, \textit{Speech} and \textit{Garbage},
for each frame. Training targets are assigned based on lattices 
produced by  Gaussian Mixture Model (GMM) based acoustic models and predefined lists
of phones for each class. GMM is used for forced alignment
as well as for unconstrained decoding. Training targets
are obtained from both procedures separately and consequently
merged by weighted summing, while samples with high disagreement
between two methods are discarded.
During the decoding, 3 estimated probabilities are transformed
to pseudo-likelihoods of 2 states, \textit{Silence} and \textit{Speech},
using priors of 3 classes and manually chosen proportions of 2 states in 3 classes.
Decoding is performed with Viterbi algorithm \cite{viterbi1967error}.

\subsection{End-to-end ASR}

End-to-end approach implements ASR system as
a single neural network based model that takes a $T$-length sequence of
$d$ dimensional feature vectors $X = \{ x_t \in \mathbb{R}^d | t = 1, \dots, T \}$ in the input
and provides a $U$-length sequence
of output labels $Y = \{ y_u \in \mathcal{U} | u = 1, \dots, U \}$,
where $\mathcal{U}$ is a set of distinct output labels
and usually $U < T$.
Common architecture for such models is attention-based encoder-decoder network
trained to minimize cross-entropy loss:
\begin{align}
\mathcal{L}_{\text{att}} = & - \log \mathnormal{p}_{\text{att}}(Y|X) \\
\mathnormal{p}_{\text{att}}(Y|X)  = & \prod_u \mathnormal{p}(y_u | X, y_{1:u-1}) \\
\mathnormal{p}(y_u | X, y_{1:u-1}) = & \text{ Decoder} (\mathbf{r}_{u}, \mathbf{q}_{u-1}, y_{u-1}) \\
\mathbf{h}_t = & \text{ Encoder}(X) \\
a_{ut} = & \text{ Attention}(\{a_{u-1}\}_t, \mathbf{q}_{u-1}, \mathbf{h}_t) \\
\mathbf{r}_u = & \sum_t a_{ut} \mathbf{h}_{t}.
\end{align}
Here, $\text{Encoder}(\cdot)$ and $\text{Decoder}(\cdot)$ are
recurrent neural  networks, $\text{Attention}(\cdot)$
is an attention mechanism and
$\mathbf{h}_t$, $\mathbf{q}_{u-1}$ and $\mathbf{r}_u$
are the hidden vectors.
Attention mechanism has been developed in the context of machine translation
problem \cite{bahdanau2014neural} and provides a means to model correspondence of all elements
of hidden representations sequence to all elements of output sequence
in the decoder. Attention mechanism allows to learn non-sequential
mapping between its inputs and outputs, meaning that
order of output elements is not always the same as order
of input elements corresponding to them,
what can be an advantage in case of machine translation task,
as word order sometimes differs between languages.
However, this property of attention mechanism makes training
of speech recognition suboptimal, because it is known in advance
that word order is the same in audio and in transcription.
Connectionist Temporal Classification (CTC) sequence level loss function \cite{graves2006connectionist} has been adopted as a secondary learning
objective for end-to-end ASR models in order to suppress this drawback:
\begin{align}
\mathcal{L} = \lambda \mathcal{L}_{\text{ctc}} + (1-\lambda) \mathcal{L}_{\text{att}},
\end{align}
where $0 \leq \lambda \leq 1$.
Encoder output followed by a single linear layer serves as estimated
output label sequence in CTC loss calculation, while target is set to be
all possible $T$-length sequences of an extended output labels set
$Z = \{ z_t \in \mathcal{U} \cup \texttt{<blank>} | t = 1, \dots, T \}$,
corresponding to the original output labels sequence $Y$:
\begin{align}
\mathcal{L}_{\text{ctc}} = & - \log \mathnormal{p}_{\text{ctc}}(Y|X) \\
\mathnormal{p}_{\text{ctc}}(Y|X)  \triangleq & \sum_Z \prod_t \mathnormal{p}(z_t | z_{t-1}, Y) \mathnormal{p}(z_t | X) \\
\mathnormal{p}(z_t | X) = & \text{ Softmax}(\text{Lin}(\mathbf{h}_t) ).
\end{align}
It has been found that CTC output can also improve decoding
results when combined with the main attention-based probabilities
during the search:
\begin{align}
\hat{Y} = \argmax_Y \{ \lambda \log \mathnormal{p}_{\text{ctc}}(Y|X) + (1-\lambda) \log \mathnormal{p}_{\text{att}}(Y|X) \}.
\end{align}

External language model (LM) is commonly-used
technique to improve ASR results. LMs are trained on text corpora,
which usually contain order of magnitude more examples of written language
compared to acoustic corpora, and therefore provide a reliable source of
information for selection of well formed transcriptions from hypotheses.
In end-to-end ASR, this information is used during
the decoding by adding LM probability of hypothetical output label sequence
with scaling factor $\gamma$
to probabilities obtained from the main model:
\begin{align}
\hat{Y} = \argmax_Y \{ \lambda \log \mathnormal{p}_{\text{ctc}}(Y|X) + 
    (1-\lambda) \log \mathnormal{p}_{\text{att}}(Y|X) + \gamma \log \mathnormal{p}_{\text{lm}}(Y) \}.
\end{align}

Encoder-decoder architecture allows output sequence (transcription)
to have any length that does not exceed length of input sequence (audio recording).
Consequently, it is possible to employ different kinds of output units,
for example words or characters. In case of words,
transcription hypotheses are limited by words presenting in vocabulary,
what causes out of vocabulary problems and requires
large dimensionality of final layers.
In case of characters, output sequences become very long
for alphabetical languages, what leads to
high number of hypothetical transcriptions
and slows down the decoding.
Sub-word units have been suggested first as a trade-off solution in
machine translation \cite{sennrich2015neural} and recently have been adopted in speech recognition \cite{zeyer2018improved}.
Sub-word units include single characters and can be used to encode
any word. In addition to that, sub-word units include
combinations of several characters and encode
words to shorter sequences compared to single characters.
Unigram language model algorithm \cite{kudo2018subword}
performs segmentation of a string $X$ by searching for the most
probable sequence of sub-word units composing the string:
\begin{align}
\mathbf{x}^{*} = \argmax_{\mathbf{x} \in \mathcal{S}(X)} P(\mathbf{x}),
\end{align}
where probability $P(\mathbf{x})$ of a sequence of sub-word units
$\mathbf{x} = (x_1,\ldots,x_M)$
is defined as the product of occurrence probabilities of sub-word
units:
\begin{align}
  P(\mathbf{x}) = \prod_{i=1}^{M} p(x_i), \\
  \forall i\,\, x_i \in \mathcal{V},\,\,\,
  \sum_{x \in \mathcal{V}} p(x) = 1. \nonumber
\end{align}
Sub-word units vocabulary $\mathcal{V}$ is derived during
the training of segmentation model by starting
from some large set of frequent in the training data
substrings and iterative
elimination of certain percent of substrings having
lowest impact on total likelihood of all possible
sequences of sub-word units for all sentences 
until some predefined size of vocabulary is reached.

\section{Implementation}

The frontend is implemented as a Node.js/React application and utilizes WebSocket protocol
to communicate with the backend. Users can sign in and upload their recordings for transcription. 
We plan to add the users' feedback with the main focus on customization and fine tuning. 

Speech segmentation is performed with Kaldi toolkit \cite{povey2011kaldi}.
We use the pretrained SAD model downloaded from \url{http://kaldi-asr.org/models/m4}.
The model is trained on Fisher-English corpus \cite{cieri2004fisher}
augmented with room impulses and additive noise from
Room Impulse Response and Noise Database \cite{ko2017study}.
The input features of SAD model
are 40-dimensional Mel Frequency Cepstral Coefficients (MFCC) without cepstral truncation
with a frame length 25 ms and shift of 10 ms.
We use the segmentation parameters suggested in 
\texttt{aspire} Kaldi recipe, but extend maximum speech segment duration
from 10 to 30 seconds and enable consecutive speech segments merging
when duration of merged segment does not exceed 10 seconds.

Speech recognition is implemented with ESPnet end-to-end
speech recognition toolkit \cite{watanabe2018espnet}
with PyTorch backend.
We follow LibriSpeech ESPnet recipe and use 
80-dimensional log Mel filterbank coefficients concatenated with
3-dimensional pitch having a frame length of 25 ms and shift of 10 ms as acoustic features
and sub-word units as output labels. 
Kaldi toolkit is used to extract and normalize input features.
Normalization
to zero mean and unit variance is done with
global statistics from the training data.
SentencePiece unsupervised text tokenizer\footnote{\url{https://github.com/google/sentencepiece}} is used to
generate list of sub-word units based on the
language model training data and to segment all kinds of text data.
We evaluated several sizes of sub-word unit vocabulary
between 50 and 5000 and found that 100 resulted in better
results for both English and German systems.
The ASR model is an encoder-decoder neural network.
The encoder network consists of 2 VGG \cite{simonyan2014very} blocks followed by
5 Bidirectional Long Short-Term Memory Network (BLSTM) layers \cite{graves2005bidirectional} with 1024 units in each layer and direction.
The decoder network consists of 2 Long Short-Term Memory Network (LSTM) \cite{hochreiter1997long} layers with 1024 units
and location based attention mechanism with 1024 dimensions,
10 convolution channels and 100 convolution filters.
CTC weight $\lambda$ is set to $0.5$ for both training
and decoding.
Training is performed with AdaDelta optimizer \cite{zeiler2012adadelta}
and gradient clipping on 4 Graphics Processing Units (GPUs) in parallel with a batch size of 24 for 10 epochs.
The optimizer is initialized with $\rho = 0.95$ and $\epsilon = 10^{-8}$.
$\epsilon$ is halved after an epoch if performance of the model
did not improve on validation set.
The model with the highest accuracy on validation set
is used for the decoding with beam size of 20. 

External LM for the English system contains 2 layers of 650 LSTM units
and is trained with stochastic gradient descent optimizer with
batch size 256 for 60 epochs.
LM scaling factor $\gamma$ is set to $0.5$
during decoding for the English system.
External LM for the German system contains 2 layers of 3000 LSTM units
and is trained with Adam optimizer \cite{kingma2014adam} with
batch size 128 for 10 epochs.
LM scaling factor $\gamma$ is set to $1.1$
during decoding for the German system.

\section{Resources}
Both English and German systems
are trained on multiple speech databases, which are summarized in
Table \ref{tab:data}.
We use data preparation scripts from \texttt{multi\_en} Kaldi recipe
and German ASR recipe \cite{milde-koehn-18-german-asr}.
German system is additionally improved by data augmentation,
applied to 3 datasets (marked with (*) in the table)
with Acoustic Simulator\footnote{\url{https://github.com/idiap/acoustic-simulator}} package.
This procedure gives an augmented dataset that is 10 times larger than original dataset.

External LM for the English system is trained with on transcriptions from the training speech databases
except of Common Voice. External LM for the German system is trained on all transcriptions
form the training speech databases and additional text
corpus\footnote{\url{http://ltdata1.informatik.uni-hamburg.de/kaldi_tuda_de/German_sentences_8mil_filtered_maryfied.txt.gz}}
containing 8 millions of preprocessed read sentences
from the German Wikipedia, the European Parliament Proceedings Parallel Corpus and a crawled corpus of direct speech.

\begin{savenotes}
\begin{table}[H]
 \centering
 \caption{English and German training data covering data sets with different styles}
  \footnotesize
  \begin{tabular}{|l|l|l|l|}
    \hline
    \textbf{Language} & \textbf{Corpus} & \textbf{Style} & \textbf{Hours} \\
    \hline
    \multirow[t]{7}{*}{English} & LibriSpeech \cite{panayotov2015librispeech} & Read & 960 \\
    \cline{2-4}
    & Switchboard \cite{godfrey1992switchboard} & Spontaneous & 317 \\
    \cline{2-4}
    & TED-LIUM 3 \cite{hernandez2018ted} & Spontaneous & 450 \\
    \cline{2-4}
    & AMI \cite{carletta2007unleashing} & Spontaneous & 229 \\
    \cline{2-4}
    & WSJ \cite{paul1992design} & Read & 81 \\
    \cline{2-4}
    & Common Voice\footnote{\url{https://voice.mozilla.org/en/datasets}} & Read & 240 \\
    \cline{2-4}
    & \textit{Total} & & \textit{2277} \\
    \hline
    \multirow[t]{8}{*}{German} & Tuda-De \cite{radeck2015open} & Read & 109 \\
    \cline{2-4}
    & SWC \cite{kohn2016mining} & Read & 245 \\
    \cline{2-4}
    & M-AILABS\footnote{\url{http://www.m-ailabs.bayern/en/the-mailabs-speech-dataset/}} (*) & Read & 2336 \\
    \cline{2-4}
    & Verbmobil 1 and 2 \cite{wahlster2013verbmobil} (*) & Mixed & 417 \\
    \cline{2-4}
    & VoxForge\footnote{\url{http://www.voxforge.org/de/Downloads}} (*) & Read & 571 \\
    \cline{2-4}
    & RVG 1 \cite{burger1998rvg} & Mixed & 100 \\
    \cline{2-4}
    & PhonDat 1 \cite{hess1995phondat} & Mixed & 19 \\
    \cline{2-4}
    & \textit{Total} & & \textit{3797} \\
    \hline
  \end{tabular}
  \label{tab:data}
\end{table}
\end{savenotes}

\section{ASR Performance}
\subsection{Results}
Table \ref{tab:results}
compares the results of IMS-Speech on several testing datasets with
the best results for the corresponding datasets which we could
find in various sources.
In summary, these results suggest that our generic systems
can compete with task specific systems and in some cases even outperform them,
possibly due to better generalization from larger amount of training data.

\begin{savenotes}
\begin{table}[H]
 \centering
  \caption{ASR performance comparison with state of the art results (WER, \%)}
  \footnotesize
  \begin{tabular}{|l|l|l|l|}
    \hline
    \textbf{Language} & \textbf{Dataset} & \textbf{IMS-Speech} & \textbf{State of the art} \\
    \hline
    \multirow[t]{7}{*}{English} & WSJ eval'92 & 3.8 & 3.5 \cite{chan2015deep} \\
    \cline{2-4}
    & LibriSpeech test-clean & 4.4 & 3.2 \cite{han2017capio} \\
    \cline{2-4}
    & LibriSpeech test-other & 12.7 & 7.6 \cite{han2017capio} \\
    \cline{2-4}
    & TED-LIUM 3 test & 12.8 & 6.7 \cite{hernandez2018ted} \\
    \cline{2-4}
    & AMI IHM eval & 17.4 & 19.2\footnote{\url{https://github.com/kaldi-asr/kaldi/blob/4bdb05ae78a842a07cae326aeb32aea87328fb2c/egs/ami/s5b/RESULTS_ihm\#L87}} \\
    \cline{2-4}
    & AMI SDM eval & 38.5 & 36.7\footnote{\url{https://github.com/kaldi-asr/kaldi/blob/4bdb05ae78a842a07cae326aeb32aea87328fb2c/egs/ami/s5b/RESULTS_sdm\#L105}} \\
    \cline{2-4}
    & AMI MDM eval & 34.1 & 34.2\footnote{\url{https://github.com/kaldi-asr/kaldi/blob/4bdb05ae78a842a07cae326aeb32aea87328fb2c/egs/ami/s5b/RESULTS_mdm\#L97}} \\
    \hline
    \multirow[t]{4}{*}{German} & Tuda-De dev & 11.1 & 13.1 \cite{milde-koehn-18-german-asr} \\
    \cline{2-4}
    & Tuda-De test & 12.0 & 14.4 \cite{milde-koehn-18-german-asr} \\
    \cline{2-4}
    & Verbmobil 1 dev & 6.7 & 18.2 \cite{milde-koehn-18-german-asr} \\
    \cline{2-4}
    & Verbmobil 1 test & 7.3 & 12.7 \cite{gaida2014comparing} \\
    \hline
  \end{tabular}
  \label{tab:results}
\end{table}
\end{savenotes}

We evaluate the recognition speech with different beam widths
and batched recognition with inference using CPU and GPU.
The results in Table \ref{tab:beam} show that batched recognition can
significantly increase speed of recognition without any impact
on WER.

\begin{table}[H]
 \centering
  \caption{Beam width effect on recognition performance and speed on Tuda-De test set}
  \footnotesize
  \begin{tabular}{|l|l|l|l|l|}
    \hline
    \multirow[t]{2}{*}{\textbf{Beam width}} &
	\multicolumn{2}{c|}{\textbf{\makecell[l]{Inference on 1 CPU core\\ with batch size of 1}}} &
	\multicolumn{2}{c|}{\textbf{\makecell[l]{Inference on 1 GPU \\with batch size of 23}}} \\\cline{2-5}
    & \textbf{WER, \%} & \textbf{RT factor} & \textbf{WER, \%} & \textbf{RT factor} \\
    \hline
    20 & 12.0 & 14.2 & 12.0 & 0.7 \\
    \hline
    15 & 12.2 & 11.3 & 12.2 & 0.5 \\
    \hline
    10 & 12.6 & 8.8 & 12.6 & 0.4 \\
    \hline
    5 & 13.7 & 7.0 & 13.7 & 0.3 \\
    \hline
  \end{tabular}
  \label{tab:beam}
\end{table}

\subsection{Comparisions with Google API}

We use the ASR benchmark framework \cite{dernoncourt2018framework} to compare
performance of IMS-Speech and Google API. The results of Google API were
retrieved on 8.01.2019. As the framework uses custom WER computation
method instead of NIST \texttt{sclite} utility used in ESPnet recipes,
we had to perform scoring of IMS-Speech output with the framework as well.
We excluded all utterances for which Google API transcriptions contained digits,
because WER would be high for them even if transcriptions were correct (a couple of examples are given in Table~\ref{tab:examples}),
and also utterances for which Google API transcriptions were empty.
The results are shown in Table \ref{tab:google}.
The numbers suggest that that Google API models may be optimized for certain speech domain
and recording conditions that differ significantly from
the ones tested by us.

\begin{table}[H]
 \centering
  \caption{Examples of some perfect IMS-Speech transcriptions and Google API transcriptions}
  \footnotesize
  \begin{tabular}{|l|l|l|}
    \hline
    \textbf{Utterance} & \textbf{System} & \textbf{Transcription} \\
    \hline
    \multirow[t]{2}{*}{\makecell[l]{LibriSpeech test-other,\\ 2609-157645-0010}} &
	IMS-Speech &
	\texttt{\makecell[l]{then let them sing to the hundred\\ and nineteenth replied the curate}} \\
    \cline{2-3} &
	Google API &
	\texttt{\makecell[l]{then let them sing the 119th\\ repository}} \\
    \hline
    \multirow[t]{2}{*}{\makecell[l]{Verbmobil 1 test,\\ w007dxx0\_001\_BFG}} &
	IMS-Speech &
	\texttt{\makecell[l]{Ich würde Ihnen den einundzwanzigsten August\\ bis zum vier fünfundzwanzigsten vorschlagen}} \\
    \cline{2-3} &
	Google API &
	\texttt{\makecell[l]{ich würde Ihnen den 21. August\\ bis den 425 vorschlagen}} \\
    \hline
  \end{tabular}
  \label{tab:examples}
\end{table}

\begin{table}[H]
\centering
  \caption{ASR performance comparison with Google API in term of WER (\%)}
  \footnotesize
  \begin{tabular}{|l|l|l|l|l|}
    \hline
    \textbf{Language} & \textbf{Dataset} & \textbf{IMS-Speech} & \textbf{Google API} & \textbf{Scored utterances} \\
    \hline
    \multirow[t]{3}{*}{English} & LibriSpeech test-clean & 4.3 & 15.9 & 2444 of 2620 (93\%) \\
    \cline{2-5} 
    & LibriSpeech test-other & 12.5  & 28.0 & 2708 of 2939 (92\%) \\
    \cline{2-5} 
    & Common Voice valid-test & 4.5 & 19.2 & 3772 of 3995 (94\%) \\
    \hline
    \multirow[t]{2}{*}{German} & Tuda-De test & 10.0 & 12.4 & 3481 of 4100 (85\%) \\
    \cline{2-5} 
    & Verbmobil 1 test &  8.7 & 19.5 & 334 of 631 (53\%) \\
    \hline
  \end{tabular}
  \label{tab:google}
\end{table}

\section{Conclusion}

We presented IMS-Speech, a web based speech transcription tool for English and German
languages that can be used by non-technical researchers
in order to utilize the information from audio recordings in their studies.
The comparison of the IMS-Speech results with the results of specialized
systems in terms of WER showed that the described service can
perform decently in a diverse set of tasks and conditions.
In the future, we plan to allow the users to customize the system
for their needs as well as to constantly improve our ASR system.

\bibliographystyle{essv}
\bibliography{paper}

\begin{thebibliography}{32}
\providecommand{\natexlab}[1]{#1}
\providecommand{\url}[1]{\texttt{#1}}
\providecommand{\urlprefix}{URL }
\expandafter\ifx\csname urlstyle\endcsname\relax
  \providecommand{\doi}[1]{doi:\discretionary{}{}{}#1}\else
  \providecommand{\doi}{doi:\discretionary{}{}{}\begingroup
  \urlstyle{rm}\Url}\fi
\input{babelbst.tex}
\newcommand{\Capitalize}[1]{\uppercase{#1}}
\newcommand{\capitalize}[1]{\expandafter\Capitalize#1}
\providecommand{\eprint}[2][]{\url{#2}}

\bibitem[Povey \bbletal{}(2011)]{povey2011kaldi}
\textsc{Povey, D.}, \textsc{A.~Ghoshal}, \textsc{G.~Boulianne},
  \textsc{L.~Burget}, \textsc{O.~Glembek}, \textsc{N.~Goel},
  \textsc{M.~Hannemann}, \textsc{P.~Motlicek}, \textsc{Y.~Qian},
  \textsc{P.~Schwarz} \textsc{\bbletal{}}: \emph{The {K}aldi speech recognition
  toolkit}.
\newblock \capitalize\bblin{} \emph{Proc. of ASRU}. 2011.

\bibitem[Watanabe \bbletal{}(2018)]{watanabe2018espnet}
\textsc{Watanabe, S.}, \textsc{T.~Hori}, \textsc{S.~Karita},
  \textsc{T.~Hayashi}, \textsc{J.~Nishitoba}, \textsc{Y.~Unno},
  \textsc{N.~E.~Y. Soplin}, \textsc{J.~Heymann}, \textsc{M.~Wiesner},
  \textsc{N.~Chen} \textsc{\bbletal{}}: \emph{{ESP}net: {E}nd-to-{E}nd {S}peech
  {P}rocessing {T}oolkit}.
\newblock \emph{arXiv preprint arXiv:1804.00015}, 2018.

\bibitem[Waibel \bbletal{}(1990)]{waibel1990phoneme}
\textsc{Waibel, A.}, \textsc{T.~Hanazawa}, \textsc{G.~Hinton},
  \textsc{K.~Shikano}, \bbland{} \textsc{K.~J. Lang}: \emph{Phoneme recognition
  using time-delay neural networks}.
\newblock \capitalize\bblin{} \emph{Readings in speech recognition}. 1990.

\bibitem[Ghahremani \bbletal{}(2016)]{ghahremani2016acoustic}
\textsc{Ghahremani, P.}, \textsc{V.~Manohar}, \textsc{D.~Povey}, \bbland{}
  \textsc{S.~Khudanpur}: \emph{{A}coustic {M}odelling from the {S}ignal
  {D}omain {U}sing {CNN}s}.
\newblock \capitalize\bblin{} \emph{Proc. of Interspeech}. 2016.

\bibitem[Viterbi(1967)]{viterbi1967error}
\textsc{Viterbi, A.}: \emph{Error bounds for convolutional codes and an
  asymptotically optimum decoding algorithm}.
\newblock \emph{IEEE Transactions on Information Theory}, 1967.

\bibitem[Bahdanau \bbletal{}(2014)]{bahdanau2014neural}
\textsc{Bahdanau, D.}, \textsc{K.~Cho}, \bbland{} \textsc{Y.~Bengio}:
  \emph{Neural machine translation by jointly learning to align and translate}.
\newblock \emph{arXiv preprint arXiv:1409.0473}, 2014.

\bibitem[Graves \bbletal{}(2006)]{graves2006connectionist}
\textsc{Graves, A.}, \textsc{S.~Fern{\'a}ndez}, \textsc{F.~Gomez}, \bbland{}
  \textsc{J.~Schmidhuber}: \emph{Connectionist temporal classification:
  labelling unsegmented sequence data with recurrent neural networks}.
\newblock \capitalize\bblin{} \emph{Proc. of ICML}. 2006.

\bibitem[Sennrich \bbletal{}(2015)]{sennrich2015neural}
\textsc{Sennrich, R.}, \textsc{B.~Haddow}, \bbland{} \textsc{A.~Birch}:
  \emph{Neural machine translation of rare words with subword units}.
\newblock \emph{arXiv preprint arXiv:1508.07909}, 2015.

\bibitem[Zeyer \bbletal{}(2018)]{zeyer2018improved}
\textsc{Zeyer, A.}, \textsc{K.~Irie}, \textsc{R.~Schl{\"u}ter}, \bbland{}
  \textsc{H.~Ney}: \emph{Improved training of end-to-end attention models for
  speech recognition}.
\newblock \emph{arXiv preprint arXiv:1805.03294}, 2018.

\bibitem[Kudo(2018)]{kudo2018subword}
\textsc{Kudo, T.}: \emph{Subword {R}egularization: {I}mproving {N}eural
  {N}etwork {T}ranslation {M}odels with {M}ultiple {S}ubword {C}andidates}.
\newblock \emph{arXiv preprint arXiv:1804.10959}, 2018.

\bibitem[Cieri \bbletal{}(2004)]{cieri2004fisher}
\textsc{Cieri, C.}, \textsc{D.~Miller}, \bbland{} \textsc{K.~Walker}: \emph{The
  {F}isher {C}orpus: a {R}esource for the {N}ext {G}enerations of
  {S}peech-to-{T}ext}.
\newblock \capitalize\bblin{} \emph{LREC}. 2004.

\bibitem[Ko \bbletal{}(2017)]{ko2017study}
\textsc{Ko, T.}, \textsc{V.~Peddinti}, \textsc{D.~Povey}, \textsc{M.~L.
  Seltzer}, \bbland{} \textsc{S.~Khudanpur}: \emph{A study on data augmentation
  of reverberant speech for robust speech recognition}.
\newblock \capitalize\bblin{} \emph{Proc. of IEEE ICASSP}. 2017.

\bibitem[Simonyan \bbland{} Zisserman(2014)]{simonyan2014very}
\textsc{Simonyan, K.} \bbland{} \textsc{A.~Zisserman}: \emph{Very deep
  convolutional networks for large-scale image recognition}.
\newblock \emph{arXiv preprint arXiv:1409.1556}, 2014.

\bibitem[Graves \bbletal{}(2005)]{graves2005bidirectional}
\textsc{Graves, A.}, \textsc{S.~Fern{\'a}ndez}, \bbland{}
  \textsc{J.~Schmidhuber}: \emph{Bidirectional {LSTM} networks for improved
  phoneme classification and recognition}.
\newblock \capitalize\bblin{} \emph{International Conference on Artificial
  Neural Networks}, \bblpp{} 799--804. Springer, 2005.

\bibitem[Hochreiter \bbland{} Schmidhuber(1997)]{hochreiter1997long}
\textsc{Hochreiter, S.} \bbland{} \textsc{J.~Schmidhuber}: \emph{Long
  short-term memory}.
\newblock \emph{Neural computation}, 9(8), \bblpp{} 1735--1780, 1997.

\bibitem[Zeiler(2012)]{zeiler2012adadelta}
\textsc{Zeiler, M.~D.}: \emph{{ADADELTA}: an adaptive learning rate method}.
\newblock \emph{arXiv preprint arXiv:1212.5701}, 2012.

\bibitem[Kingma \bbland{} Ba(2014)]{kingma2014adam}
\textsc{Kingma, D.~P.} \bbland{} \textsc{J.~Ba}: \emph{Adam: {A} method for
  stochastic optimization}.
\newblock \emph{arXiv preprint arXiv:1412.6980}, 2014.

\bibitem[Milde \bbland{} K{\"o}hn(2018)]{milde-koehn-18-german-asr}
\textsc{Milde, B.} \bbland{} \textsc{A.~K{\"o}hn}: \emph{{O}pen {S}ource
  {A}utomatic {S}peech {R}ecognition for {G}erman}.
\newblock \capitalize\bblin{} \emph{Proc. of ITG}. 2018.

\bibitem[Panayotov \bbletal{}(2015)]{panayotov2015librispeech}
\textsc{Panayotov, V.}, \textsc{G.~Chen}, \textsc{D.~Povey}, \bbland{}
  \textsc{S.~Khudanpur}: \emph{Librispeech: an {ASR} corpus based on public
  domain audio books}.
\newblock \capitalize\bblin{} \emph{Proc. of IEEE ICASSP}. 2015.

\bibitem[Godfrey \bbletal{}(1992)]{godfrey1992switchboard}
\textsc{Godfrey, J.~J.}, \textsc{E.~C. Holliman}, \bbland{}
  \textsc{J.~McDaniel}: \emph{{SWITCHBOARD}: {T}elephone speech corpus for
  research and development}.
\newblock \capitalize\bblin{} \emph{Proc. of IEEE ICASSP}. 1992.

\bibitem[Hernandez \bbletal{}(2018)]{hernandez2018ted}
\textsc{Hernandez, F.}, \textsc{V.~Nguyen}, \textsc{S.~Ghannay},
  \textsc{N.~Tomashenko}, \bbland{} \textsc{Y.~Esteve}: \emph{{TED}-{LIUM} 3:
  twice as much data and corpus repartition for experiments on speaker
  adaptation}.
\newblock \emph{arXiv preprint arXiv:1805.04699}, 2018.

\bibitem[Carletta(2007)]{carletta2007unleashing}
\textsc{Carletta, J.}: \emph{Unleashing the killer corpus: experiences in
  creating the multi-everything {AMI} {M}eeting {C}orpus}.
\newblock \emph{Language Resources and Evaluation}, 2007.

\bibitem[Paul \bbland{} Baker(1992)]{paul1992design}
\textsc{Paul, D.~B.} \bbland{} \textsc{J.~M. Baker}: \emph{The design for the
  {W}all {S}treet {J}ournal-based {CSR} corpus}.
\newblock \capitalize\bblin{} \emph{Proc. of the workshop on Speech and Natural
  Language}. 1992.

\bibitem[Radeck-Arneth \bbletal{}(2015)]{radeck2015open}
\textsc{Radeck-Arneth, S.}, \textsc{B.~Milde}, \textsc{A.~Lange},
  \textsc{E.~Gouv{\^e}a}, \textsc{S.~Radomski}, \textsc{M.~M{\"u}hlh{\"a}user},
  \bbland{} \textsc{C.~Biemann}: \emph{Open source german distant speech
  recognition: {C}orpus and acoustic model}.
\newblock \capitalize\bblin{} \emph{Text, Speech, and Dialogue}. 2015.

\bibitem[K{\"o}hn \bbletal{}(2016)]{kohn2016mining}
\textsc{K{\"o}hn, A.}, \textsc{F.~Stegen}, \bbland{} \textsc{T.~Baumann}:
  \emph{Mining the {S}poken {W}ikipedia for {S}peech {D}ata and {B}eyond}.
\newblock \capitalize\bblin{} \emph{Proc. of LREC}. 2016.

\bibitem[Wahlster(2013)]{wahlster2013verbmobil}
\textsc{Wahlster, W.}: \emph{Verbmobil: foundations of speech-to-speech
  translation}.
\newblock Springer Science \& Business Media, 2013.

\bibitem[Burger \bbland{} Schiel(1998)]{burger1998rvg}
\textsc{Burger, S.} \bbland{} \textsc{F.~Schiel}: \emph{{RVG} 1 -- {A}
  {D}atabase for {R}egional {V}ariants of {C}ontemporary {G}erman}.
\newblock \capitalize\bblin{} \emph{Proc. of LREC}. 1998.

\bibitem[Hess \bbletal{}(1995)]{hess1995phondat}
\textsc{Hess, W.~J.}, \textsc{K.~J. Kohler}, \bbland{} \textsc{H.-G. Tillmann}:
  \emph{The {P}hondat-verbmobil speech corpus}.
\newblock \capitalize\bblin{} \emph{Fourth European Conference on Speech
  Communication and Technology}. 1995.

\bibitem[Chan \bbland{} Lane(2015)]{chan2015deep}
\textsc{Chan, W.} \bbland{} \textsc{I.~Lane}: \emph{Deep recurrent neural
  networks for acoustic modelling}.
\newblock \emph{arXiv preprint arXiv:1504.01482}, 2015.

\bibitem[Han \bbletal{}(2017)]{han2017capio}
\textsc{Han, K.~J.}, \textsc{A.~Chandrashekaran}, \textsc{J.~Kim}, \bbland{}
  \textsc{I.~Lane}: \emph{The {CAPIO} 2017 conversational speech recognition
  system}.
\newblock \emph{arXiv preprint arXiv:1801.00059}, 2017.

\bibitem[Gaida \bbletal{}(2014)]{gaida2014comparing}
\textsc{Gaida, C.}, \textsc{P.~Lange}, \textsc{R.~Petrick}, \textsc{P.~Proba},
  \textsc{A.~Malatawy}, \bbland{} \textsc{D.~Suendermann-Oeft}: \emph{Comparing
  open-source speech recognition toolkits}.
\newblock \emph{Tech. Rep., DHBW Stuttgart}, 2014.

\bibitem[Dernoncourt \bbletal{}(2018)]{dernoncourt2018framework}
\textsc{Dernoncourt, F.}, \textsc{T.~Bui}, \bbland{} \textsc{W.~Chang}: \emph{A
  {F}ramework for {S}peech {R}ecognition {B}enchmarking}.
\newblock \emph{Proc. of Interspeech}, 2018.

\end{thebibliography}

\end{document}